\documentclass[10pt,journal,compsoc]{IEEEtran}
\ifCLASSOPTIONcompsoc
  \usepackage[nocompress]{cite}
\else
  \usepackage{cite}
\fi
\ifCLASSINFOpdf
  \usepackage[pdftex]{graphicx}
  \graphicspath{{../figures/}}
  \DeclareGraphicsExtensions{.pdf,.jpeg,.png}
\else
\fi
\usepackage{amsmath}
\usepackage{array}

\ifCLASSOPTIONcompsoc
 \usepackage[caption=false,font=footnotesize,labelfont=sf,textfont=sf]{subfig}
\else
 \usepackage[caption=false,font=footnotesize]{subfig}
\fi
\usepackage{fixltx2e}

\usepackage{stfloats}
\usepackage{url}

\usepackage{amssymb}
\usepackage{makecell}
\usepackage{pifont}
\usepackage{colortbl}
\usepackage{booktabs}
\usepackage{hyperref}
\hypersetup{
    colorlinks=true,
    linkcolor=black,
    filecolor=black,      
    urlcolor=cyan,
    citecolor=black,
    }

\definecolor{demphcolor}{RGB}{100,100,100}
\newcommand{\demph}[1]{\textcolor{demphcolor}{#1}}
\newcommand{\xmark}{\ding{55}}
\newcommand{\cmark}{\ding{51}}
\newcommand{\mypm}[1]{{\tiny{{\demph{{$\pm$#1}}}}}}
\newcommand{\smallmul}[1]{{\tiny{{{$\times$#1}}}}}

\newcolumntype{?}{!{\vrule width 1.5pt}}

\definecolor{changeclr}{RGB}{0,0,0}
\newcommand{\change}[1]{\textcolor{changeclr}{#1}}

\definecolor{c1}{RGB}{0,0,0}
\definecolor{c2}{RGB}{ 31, 119, 180}
\definecolor{c3}{RGB}{255, 127,  14}
\definecolor{c4}{RGB}{ 44, 160,  44}
\definecolor{c5}{RGB}{214,  38,  39}
\definecolor{c6}{RGB}{148, 102, 189}
\definecolor{c7}{RGB}{174, 199, 232}

\newcommand{\mybox}[1]{\textcolor{#1}{\rule{0.7em}{0.7em}}}

\begin{document}
%
\title{Fast Weakly Supervised Action Segmentation Using Mutual Consistency}
%
%
%
%

\author{Yaser~Souri$^\ast$, 
        Mohsen~Fayyaz$^\ast$, 
        Luca~Minciullo, 
        Gianpiero~Francesca, 
        and~Juergen~Gall,~\IEEEmembership{Member,~IEEE}
\IEEEcompsocitemizethanks{\IEEEcompsocthanksitem Y. Souri, M. Fayyaz and J. Gall are with the University of Bonn, Germany.
L. Minciullo and G. Francesca are with Toyota Motor Europe.

$\ast$ indicates equal contribution.

E-mail: \{souri, fayyaz, gall\}@iai.uni-bonn.de
}
\thanks{Manuscript received April 19, 2005; revised August 26, 2015.}}

%
%

\markboth{Journal of \LaTeX\ Class Files,~Vol.~14, No.~8, August~2015}%
{Souri \MakeLowercase{\textit{et al.}}: Fast Weakly Supervised Action Segmentation Using Mutual Consistency}
%



\IEEEtitleabstractindextext{%
\begin{abstract}
Action segmentation is the task of predicting the actions for each frame of a video. As obtaining the full annotation of videos for action segmentation is expensive, weakly supervised approaches that can learn only from transcripts are appealing. In this paper, we propose a novel end-to-end approach for weakly supervised action segmentation based on a two-branch neural network. The two branches of our network predict two redundant but different representations for action segmentation and we propose a novel mutual consistency (MuCon) loss that enforces the consistency of the two redundant representations. 
Using the MuCon loss together with a loss for transcript prediction, our proposed approach achieves the accuracy of state-of-the-art approaches while being $14$ times faster to train and $20$ times faster during inference.
The MuCon loss proves beneficial even in the fully supervised setting.
\end{abstract}

}

\maketitle

\IEEEdisplaynontitleabstractindextext

%
\IEEEpeerreviewmaketitle

\IEEEraisesectionheading{\section{Introduction}\label{sec:introduction}}

%
%
%
%
\IEEEPARstart{T}{he} production, availability, and consumption of video data is increasing every day at an exponential rate. With such growth comes the need to analyze, monitor, annotate, and learn from this huge amount of often publicly available data. The research community has therefore shown great interest in video data and approaches for action recognition on trimmed video clips have shown remarkable results in recent years~\cite{twostream,i3d,fayyazetal,T3D,HVU,nonlocal,slowfast}.
Although these results on trimmed video clips are important,
in many real-world scenarios we are often faced with untrimmed videos that contain multiple actions with different lengths~\cite{kuehne2016end,MS-TCN,lea2017temporal,zhao2017temporal}.


Since annotating exact temporal boundaries of actions in long videos is cumbersome and costly, action segmentation approaches that can utilize weaker types of supervision are important.
\change{In particular ordered lists of actions such as \emph{spoon powder - pour milk - stir milk}, which are termed transcripts and describe the order of the actions as they occur in a training video but not when they occur, are a popular type of weak supervision~\cite{ectc,richard2017weakly,richard2018nnviterbi,isba,d3tw,CDFL}}.  
To learn from transcripts,
previous approaches try to align the transcripts to the training videos, \emph{i.e}., they infer frame-wise labels of each training video based on the provided transcripts.
This alignment is then used as the pseudo ground truth for training.
For the transcript alignment, the Viterbi algorithm is commonly used. It takes the estimated frame-wise class probabilities of a video and finds the best sequence of frame labels that does not violate the action order of the given transcript. While~\cite{richard2017weakly,isba} perform the alignment after each epoch for all training videos, \cite{richard2018nnviterbi,CDFL} apply it at each iteration to a single video.

\begin{figure}
 \centering
 \includegraphics[width=0.48\textwidth]{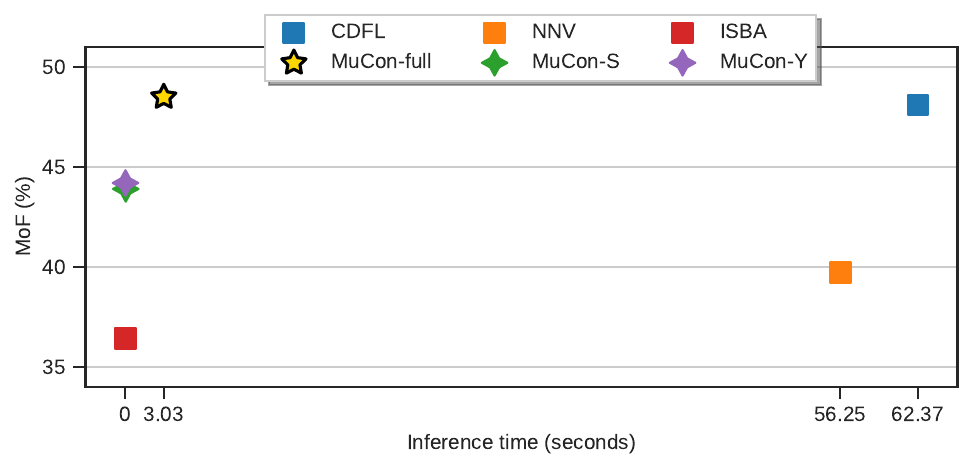}
 \vspace{-2mm}
 \caption{
 \change{
 Average inference time per video (seconds) vs.\ average mean over frames (MoF) accuracy (\%) for weakly supervised action segmentation approaches on the Breakfast dataset~\cite{breakfast}. The average MoF is calculated over 5 training/inference iterations. The proposed approach (MuCon-full) provides the best trade-off between inference time and accuracy. More details are provided in Section~\ref{sec:experiments:speed}. 
 }
 }
 \label{fig:speed_accuracy}
\end{figure}

\change{
During inference, previous approaches except ISBA~\cite{isba} rely on segmentation through alignment. This means that given an unseen video from the test set, the methods search over all transcripts of the training set and take the transcript that best aligns with the test video. This is highly undesirable since the inference time increases in this case as the number of different transcripts in the training set increases. As a result, these approaches are inefficient as shown in Figure~\ref{fig:speed_accuracy}.
}
In contrast to segmentation through alignment approaches, ISBA is fast, but it does not achieve the accuracy of state-of-the-art approaches. This means that at the moment one can only choose between accurate or fast approaches, but one cannot have both. 


In this work, we therefore fill this gap and propose an approach that is nearly as fast as ISBA and nearly as accurate as the state-of-the-art as shown in Figure~\ref{fig:speed_accuracy}. Instead of optimizing over all possible transcripts during inference, our approach directly predicts the transcript for a video as well as the frame-wise class probabilities using two branches as illustrated in Figure~\ref{fig:overview}. This means that the inference time does not depend on the number of transcripts used for training and that the best label sequence can be directly predicted from the estimated transcript and frame-wise class probabilities. 

In addition, the proposed approach has also two major advantages during training. First, the branch, which predicts the transcripts, is directly trained with the type of supervision that is provided, namely transcripts. Second, the branch not only predicts the transcripts, but also the length of each action in the transcript. This means that the branch predicts already a full segmentation of the video without the need for the Viterbi algorithm, which also reduces the training time. Nevertheless, we need an additional loss to learn the action lengths. In contrast to previous works, we do not generate pseudo ground truth with hard labels during training, but we propose a novel differentiable mutual consistency (MuCon) loss that enforces that the representations estimated by the two branches are mutually consistent and match each other. \change{Furthermore, we show that the approach can be trained using weak supervision, full supervision, or a mixture of the two, where only few videos are fully annotated.} 

We provide a thorough analysis of the proposed approach including a detailed statistical analysis. 
We show that the proposed network with the mutual consistency loss achieves an accuracy that is either on par or better than existing approaches. At the same time, it is 20 times faster during inference compared to the most accurate approach~\cite{CDFL} as shown in Figure~\ref{fig:speed_accuracy}.

\section{Related Work}\label{sec:related}
Action segmentation in untrimmed videos with various levels of supervision has been studied in several works. We thus first describe approaches that use full supervision and then approaches that use weaker levels of supervision. Finally, we discuss sequence to sequence learning methods that are related to our work.

\textbf{Fully Supervised Action Segmentation.}~~
Action segmentation in videos has already been tackled in many works \cite{spriggs2009temporal,kuehne2016end,lea2017temporal,zhao2017temporal,MS-TCN}.
Early approaches for action segmentation rely on multi-scale sliding window processing \cite{spriggs2009temporal,gaidon2013temporal,rohrbach2012database,karaman2014fast} or Markov models on top of frame-classifiers \cite{kuehne2016end,lea2016segmental}. These approaches are typically slow at inference time and do not provide strong temporal models.
More recent approaches for fully supervised action segmentation aim to capture long-range temporal dependencies using temporal convolutions \cite{MS-TCN,lea2017temporal,mstcnpp}.
\change{Some works for action segmentation focus on segmenting egocentric videos \cite{poleg2014temporal,furnari2018personal}, including also additional sensors \cite{spriggs2009temporal}.}

\textbf{Weakly Supervised Action Segmentation.}~~
Weakly supervised action segmentation has gained increased attention over the last years. Bojanowski et~al.~\cite{hollywoodextended} introduced the Hollywood extended dataset and proposed a method based on discriminative clustering for the task of action alignment.
Huang et~al.~\cite{ectc} have proposed to use an extended version of the connectionist temporal classification loss where they take the similarity of frames of the same action into account. Inspired by methods that are used in speech processing, Kuehne et~al.~\cite{hildecviu} proposed an approach based on hidden Markov models and use a Gaussian mixture model as observation model. It iteratively generates a pseudo ground truth for videos at the start of each epoch and refines the pseudo ground truth at the end of the epoch. The approaches~\cite{richard2017weakly,isba,kuehne2018hybrid} build on this work by replacing the Gaussian mixture model by a recurrent neural network for short-range temporal modeling, but the methods keep the iterative nature that involves the generation of pseudo ground truth at each epoch.

While these methods rely on iterative approaches with two-step optimization, which does not allow for direct end-to-end training, Richard et~al.~\cite{richard2018nnviterbi} introduced the Neural Network Viterbi (NNV) method. This method generates the pseudo ground truth for each iteration instead of each epoch, but it needs an additional buffer which increases the training time. They use furthermore a global length model for actions, which is updated heuristically during training. Recently, Li et~al.~\cite{CDFL} introduced an extension of NNV that outperformed existing works in terms of accuracy on standard benchmarks. They introduced a new constrained discriminative loss that discriminates between the energy of valid and invalid segmentations of the training videos. This results in a large improvement in accuracy compared to NNV.

\textbf{Sequence to Sequence Learning.}~~
There has been a lot of work in sequence to sequence learning, mostly in natural language processing \cite{seq2seq,seq2seqwithattention,convolutionalseq2seq,transformer}. However, our problem is different in one major aspect: the size mismatch between the video features and transcripts is often large. In the Breakfast dataset \cite{breakfast}, for example, the average input video length is around 2100 frames and the longest video has around 9700 frames, while the average transcript length is 6.8 and the longest is 25. Because of this issue, we have specially designed our network to be able to temporally model such long sequences. Wei et~al.~\cite{s2seg} proposed a sequence to sequence model for detecting temporal segments of interest, but not the action classes. 

\section{Weakly Supervised Action Segmentation}\label{sec:weak_seg}
Action segmentation is the task of predicting the action class for each frame of a video.
More formally, given an input sequence of $T$ $D$-dimensional frame-level features $X_{1:T} = [x_1, \dots, x_T]$, $x_t \in \mathbb{R}^D$, the goal is to predict the output sequence of frame-level action labels $\hat{Y}_{1:T}=[\hat{y}_1,\dots,\hat{y}_T]$, where $\hat{y}_{t} \in \mathcal{C}$ and $\mathcal{C}$ is the set of action classes.
The frame-level action labels $\hat{Y}_{1:T}$ can also be represented as an ordered sequence of $M$ segments $\hat{S}_{1:M}$ where each segment $\hat{s}_m$ is defined as an action label $\hat{a}_m \in \mathcal{C}$ and its corresponding length $\hat{\ell}_m \in \mathbb{R_{+}}$.

For fully supervised action segmentation, the target labels for every frame $\hat{y}_t$ are known during training. This means that the target lengths $\hat{\ell}_m$ are known.
However, in weakly supervised action segmentation, the only supervisory signal is the ordered sequence of actions $\hat{A}_{1:M}=[\hat{a}_1,\ldots,\hat{a}_m]$, often called \emph{video transcript}, while the action lengths $\hat{L}_{1:M}=[\hat{\ell}_1,\ldots,\hat{\ell}_M]$ are unknown. 

In order to estimate ${L}_{1:M}$, we exploit the fact that the two unknown target representations $\hat{Y}_{1:T}$ and $\hat{S}_{1:M}$ for action segmentation are redundant and it is possible to generate one given the other. The main idea is therefore to predict both representations ${Y}_{1:T}$ and ${S}_{1:M}$ by different branches of the network such that they can supervise each other. In the ideal case, both predicted representations converge to the same solution during training. In order to train the model, however, we have to face the challenging problem that the mapping from one representation to the other needs to be differentiable, such that the loss that measures the mutual consistency of both representations is differentiable with respect to the predicted frame-wise class probabilities ${Y}_{1:T}$ and the predicted segment lengths ${L}_{1:M}$. Since the transcript $\hat{A}_{1:M}$ is given, it does not need to be differentiable with respect to ${A}_{1:M}$.               


\section{Proposed Method}\label{sec:proposed}
\begin{figure*}[t]
 \centering
 \includegraphics[width=0.85\textwidth]{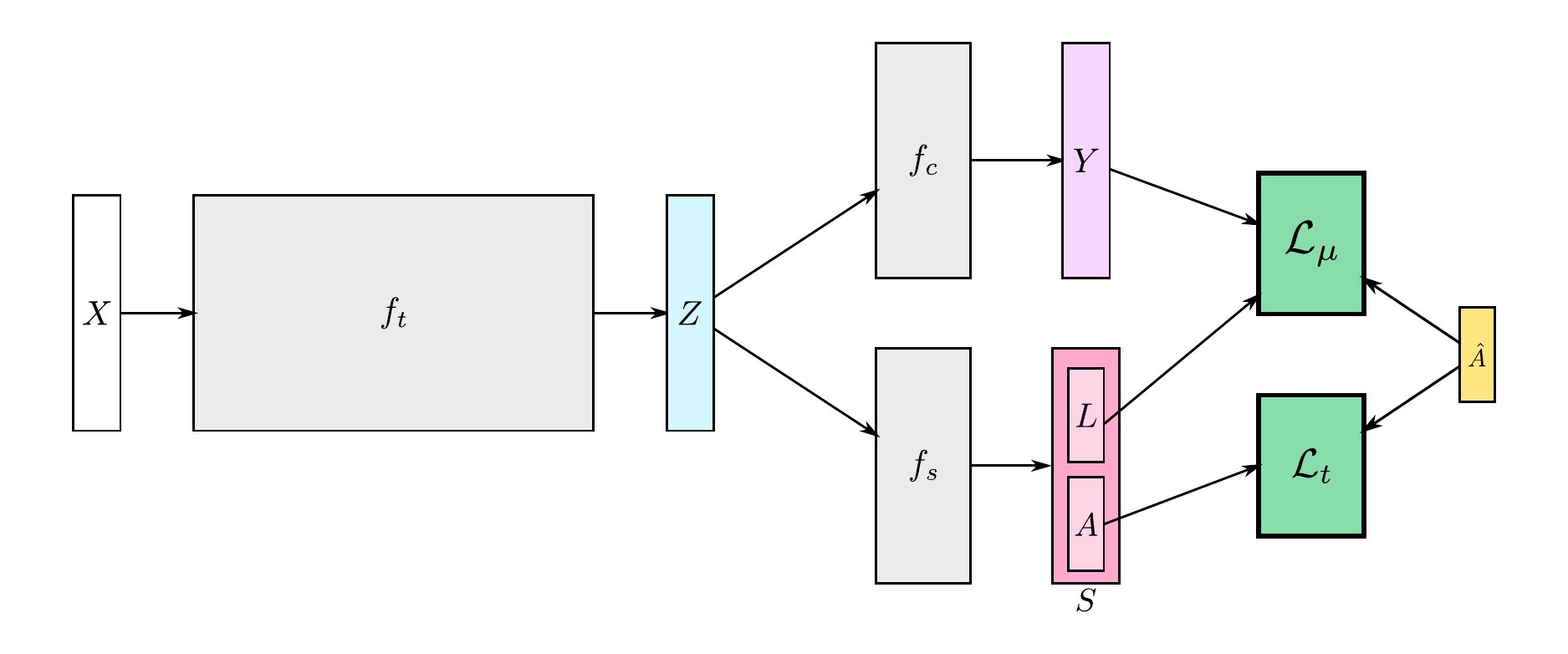}
 \vspace{-6mm}
 \caption{Our proposed network consists of three subnetworks (gray). The temporal backbone $f_t$ embeds the input features in the hidden representation $Z$ which is used for two branches.
 While the frame classification branch $f_c$ predicts framewise class probabilities $Y$ for action segmentation, the segment generation branch $f_s$ predicts the segment representation $S$ for action segmentation. We train our network using two loss functions. While the transcript prediction loss $\mathcal{L}_t$ enforces that the predicted transcript $A$ matches the ground-truth transcript $\hat{A}$, our proposed mutual consistency (MuCon) loss $\mathcal{L}_{\mu}$ enforces that the two representations are consistent.}
 \label{fig:overview}
\end{figure*}

During training, we have for each video the input video features
$X_{1:T}$ and its corresponding transcript $\hat{A}_{1:M}$.
Since only weak labels in form of the transcripts are provided for training, we propose a) to use the weak labels directly for training a transcript prediction subnetwork and b) exploit the two representations discussed in Section~\ref{sec:weak_seg} as mutual supervision.

The proposed network is illustrated in Figure~\ref{fig:overview}. The network consists of two branches that share the same backbone $f_t(X)$, which is a temporal convolutional network that maps the input video features $X \in \mathbb{R}^{T \times D}$ to a latent video representation $Z \in \mathbb{R}^{T' \times D'}$. The backbone architecture is described in Section~\ref{sec:method:temp}. After the shared backbone, the network has two branches. The top branch $f_c(Z)$ is a standard frame classification branch that estimates the class probabilities of each frame $Y \in \mathbb{R}^{T \times N}$, where $N$ is the number of classes. The branch will be described in Section~\ref{sec:method:fc}. The novelty of the network is the lower branch and the mutual consistency loss $\mathcal{L}_{\mu}$, which will be described in Sections \ref{sec:method:fs} and \ref{sec:method:mucon}, respectively.

In contrast to previous works that commonly use a network to predict $Y$ and Viterbi decoding on top of it for computing the loss based on the given transcript $\hat{A}$, we use a second branch that predicts the segment representation $S$, \emph{i.e.} $A$ and $L$. This has two advantages during training. First, we can compare the predicted transcript $A$ and the ground-truth transcript $\hat{A}$ for each video directly, \emph{i.e.}, the network is directly trained with the type of supervision that is provided. The corresponding loss is denoted by $\mathcal{L}_{t}$. Second, the network has two branches that supervise each other and we do not need an additional Viterbi decoding step. 
Since the two branches predict different representations, the novel mutual consistency (MuCon) loss $\mathcal{L}_{\mu}$ requires a differentiable mask generator such that the loss is differentiable with respect to $Y$ and $L$; otherwise we could not train the network. 

After training, we need to infer the action class for each frame of an unseen test video. In case of action segmentation, no additional transcripts are given for the test videos. As shown in Figure~\ref{fig:speed_accuracy}, we can use three different approaches for inference, which are denoted by MuCon-Y, MuCon-S, and MuCon-full. In case of MuCon-Y and MuCon-S, we use the predicted $Y$ or $S$ representation, respectively. While these settings are as fast but more accurate than ISBA~\cite{isba}, the accuracy is lower compared to the state-of-the-art. For the model MuCon-full, we use therefore the predictions of both branches. To this end, we use the predicted framewise class probabilities $Y$ and the predicted transcript $A$ to find the best sequence of action labels using Viterbi decoding as described in Section~\ref{sec:method:inference}. Note that the approach is still much faster than other methods except of ISBA since we do not need to optimize over all possible transcripts. We now provide more details for each part of the network.

\subsection{Temporal Backbone}\label{sec:method:temp}

The temporal backbone $f_t(X)$, which is a temporal convolutional network~\cite{WaveNet,MS-TCN,lea2017temporal,mstcnpp}, outputs the latent video representation $Z \in \mathbb{R}^{T' \times D'}$ of the input video features $X \in \mathbb{R}^{T \times D}$. This hidden representation has a smaller temporal resolution due to temporal pooling.
Similar to~\cite{MS-TCN,mstcnpp}, our temporal backbone consists of a set of 1-dimensional dilated convolutional layers with increasing dilation sizes. More specifically, we first apply a 1-d convolution with kernel size $1$ to perform dimensionality reduction. Then a set of 11 layers with increasing dilation rates are applied, followed by a single 1-d convolution with kernel size $1$ that generates the output. The amount of dilation for each layer is $2^i$. We perform temporal max pooling with a kernel size of $2$ after the layers 1, 2, 4, and 8. Additional details are given in Section~\ref{sec:experiments:details}. The design is similar to a single stage TCN \cite{MS-TCN}, but it includes additional pooling layers.


\subsection{Frame Classification Branch}\label{sec:method:fc}
The classification branch takes the shared latent video representation $Z \in \mathbb{R}^{T' \times D'}$ as input and predicts the class probabilities $Y \in \mathbb{R}^{T \times N}$ where $N$ is the number of actions in the dataset. Due to the temporal pooling of the temporal backbone, $Z$ has a lower temporal resolution compared to the input features. To compensate for this, we first upsample $Z$ using nearest-neighbor interpolation to the desired temporal size $T$. Then we use a single 1-d convolution with kernel size $1$, which takes as input the upsampled $Z \in \mathbb{R}^{T \times D'}$ and outputs $Y \in \mathbb{R}^{T \times N}$.

\begin{figure*}
 \centering
 \subfloat[]{
 \includegraphics[width=10cm]{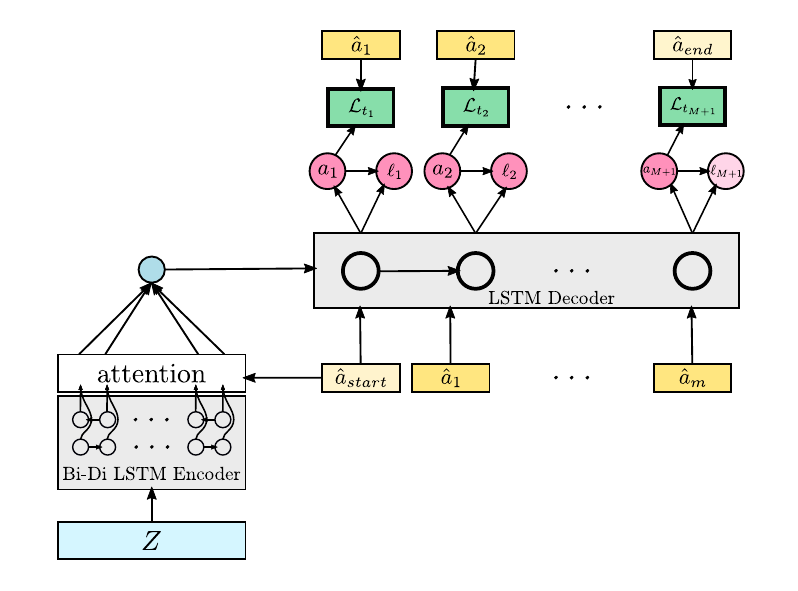}
 \label{fig:segment_gen}
 }
 \subfloat[]{
 \includegraphics[width=7.5cm]{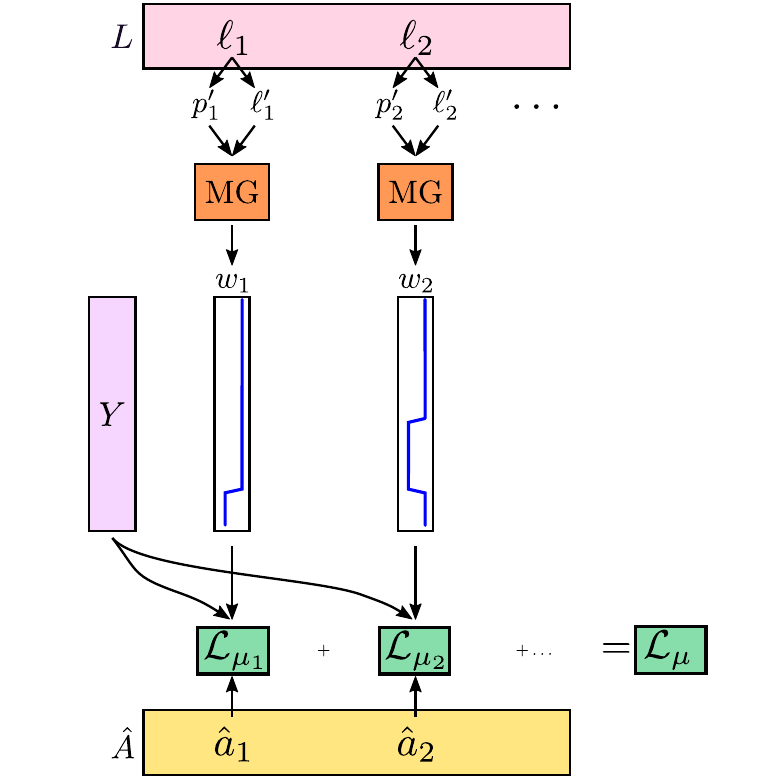}
 \label{fig:weak_loss}
 }
 \caption[Visualization of the segment generation branch and the loss functions.]{
 \change{Visualization of the segment generation branch and the loss functions.
 \textbf{\subref{fig:segment_gen}}
 Segment generation branch $f_s$ and the transcript prediction loss $\mathcal{L}_t$.
 Given the hidden video representation $Z$, we use a sequence to sequence network with attention.
 The transcript prediction loss $\mathcal{L}_t$ compares the predicted action class probabilities $a_m$ with the ground-truth action label $\hat{a}_m$.
 \textbf{\subref{fig:weak_loss}}
 Mutual consistency loss $\mathcal{L}_{\mu}$.
 Given the predicted lengths $L$, a set of masks $w_m$ are generated using differentiable sampling by the mask generation module (MG). The loss then measures for each segment the consistency of the estimated framewise class probabilities $Y$ with the ground-truth action $\hat{a}_m$.}
 }
\end{figure*}

\subsection{Segment Generation Branch}\label{sec:method:fs}
The second branch on top of $Z$ is the segment generation subnetwork which predicts the segments $S$.
Each segment $s_m$ consists of predicted action probabilities $a_m$ and the estimated relative log length $\ell_m$ of that segment. By predicting the log length, we implicitly enforce that the segment lengths are positive. 
We will discuss in Section \ref{sec:loc} how the relative log length is mapped to the absolute length.
The subnetwork and the transcript prediction loss $\mathcal{L}_t$ are illustrated in Figure~\ref{fig:segment_gen}.

We employ a conventional sequence to sequence network with attention \cite{seq2seqwithattention}.
Given the hidden video representation $Z$, we use a bidirectional LSTM encoder to encode it.
Our decoder is an LSTM recurrent neural network with MLP attention.
Although these networks on their own struggle to learn a temporal model for long input sequences~\cite{cho2014properties,Singh_2016_CVPR}, the temporal backbone, which encodes temporal relations at higher resolution, makes it easier for the segment generation subnetwork to learn temporal dependencies as shown by our ablation experiments.

As illustrated in Figure~\ref{fig:segment_gen},
the decoding starts with the starting symbol $\hat{a}_{start}$.
At each step of the decoding process, the ground truth action label $\hat{a}_{m-1}$ from the previous step is added to the encoded input sequence by concatenating it to the result of the attention. The concatenated vector is then given as input to the LSTM decoder cell, which estimates probability scores $a_m$ using a fully connected MLP with two layers. Given these probability scores and the ground truth action label $\hat{a}_m$, we compute the action prediction loss $\mathcal{L}_{t_m}$ per segment using cross-entropy.
The final transcript prediction loss is defined as the sum of the action prediction losses, \emph{i.e.}, $\mathcal{L}_{t} = \sum_{m=1}^{M+1} \mathcal{L}_{t_m}$. Note that we have $M+1$ terms since we add the end symbol $\hat{a}_{end}$ to the ground-truth transcripts, which needs to be predicted by the network as well. During training we use teacher forcing \cite{teacherforcing}, \emph{i.e.}, we do not sample from the predicted action probabilities to feed the decoder, but we use the ground truth action labels $\hat{a}_{m}$.

Given the probability scores $a_m$ and the hidden state of the decoder, we use another fully connected MLP with two layers to predict $\ell_m$, which corresponds to the logarithm of the relative length of the segment. Notice that the parameters of this MLP are not updated based on the transcript prediction loss, but based on the mutual consistency loss $\mathcal{L}_{\mu}$.





\subsection{Mutual Consistency Loss}\label{sec:method:mucon}
Using the frame classification and the segment generation branches, two representations $Y$ and $S$ for the action segmentation are produced. However, so far we have only defined a loss for the predicted transcript $A$ of the segment representation $S$, which we compare directly with the ground-truth transcript $\hat{A}$, but not for the predicted lengths of the actions $L$ and the framewise class probabilities $Y$.    
We therefore propose the mutual consistency (MuCon) loss, which enforces that the two representations match each other and are mutually consistent. As shown in Figure~\ref{fig:overview}, the mutual consistency loss takes the ground-truth transcript $\hat{A}$, the predicted segment lengths $L$, and framewise class probabilities $Y$ as input. Since the loss is used to train both branches, it needs to be differentiable with respect to $L$ and $Y$.

In principle, there are two choices: either we map $Y$ to $S$ or $S$ to $Y$. The first approach, however, is not practical since it would need to detect consistent segments within $Y$. While this could be done using Viterbi decoding, it would make our network as expensive as segmentation through alignment approaches. Our goal, however, is to propose an approach that is fast and accurate. An explicit mapping from $S$ to $Y$ followed by the computation of a framewise loss, however, is also inefficient. We therefore combine the mapping and loss computation. Furthermore, we use $\hat{A}$ instead of $A$ since we have already a loss that enforces that $A$ is close to $\hat{A}$. \change{Like teacher forcing, it also stabilizes the training since it guides the mutual consistency loss by $\hat{A}$ from the beginning of the training when $A$ is still noisy. In our experiments, we show that this reduces the standard deviation over different runs.}  

The computation of the mutual consistency loss is illustrated in Figure~\ref{fig:weak_loss}. We start with the estimated relative log length $\ell_m$ for each segment $s_m$. The relative log length $\ell_m$ is then converted into the absolute length $\ell'_m$ and for each segment we compute its absolute starting position $p'_m$ within the video. This step is described in Section \ref{sec:loc}. Given $\ell'_m$ and $p'_m$, for each segment we generate a mask $w_m$ using the differentiable mask generation module, which is described in Section \ref{sec:mask}. As mentioned before, for each segment we use the ground-truth action label $\hat{a}_m$ instead of the estimated class probabilities. 

Finally, we can compare the consistency of the predicted frame-wise class probabilities $Y$ with each segment defined by the mask $w_m$ and the label $\hat{a}_m$. To this end, we first compute the average of $Y$ for each segment based on the mask $w_m$:
\begin{equation}
 \label{eq:window_average}
 g(Y, w_m) = \frac{\sum_{t=1}^{T} y_t w_m[t]}{\ell'_m}
\end{equation}
where $g(Y, w_m) \in \mathbb{R}^N $, $w_m[t]$ is the value of the mask at frame $t$, and $\ell'_m$ is the absolute length of the segment. 
Then, we compute the cross-entropy for each segment: 
\begin{equation}
 \label{eq:mucon_m}
 \mathcal{L}_{\mu_m}(Y, w_m, \hat{a}_m) = - \log\left(\frac{e^{g(Y,w_m)[\hat{a}_m]}}{\sum_{n=1}^{N} e^{g(Y,w_m)[n]}}\right)
\end{equation}
where we use the softmax function to normalize the class probabilities per segment and the ground-truth label $\hat{a}_m$. Since we normalize per segment using the softmax function, we use the estimates of $Y$ before the softmax layer of the frame classification branch in \eqref{eq:window_average}.       
The final mutual consistency loss $\mathcal{L}_{\mu}$ is defined as the sum of all segment losses: 
\begin{equation}\label{eq:mucon}
\mathcal{L}_{\mu} = \sum_{m=1}^{M} \mathcal{L}_{\mu_m}(Y, w_m, \hat{a}_m).
\end{equation}

The mutual consistency loss \eqref{eq:mucon} is a differentiable function of the masks $w_m$ and the frame-wise class probabilities $Y$. Due to the definition of the differentiable mask generation, which we will describe in the following section, the masks are differentiable with respect to the estimated segment lengths $L$. This means that the gradients of the mutual consistency loss are backpropagated through both branches as it is required to train the network.

\subsection{Differentiable Mask Generation}\label{sec:mask}

\begin{figure}
 \centering
 \includegraphics[width=0.48\textwidth]{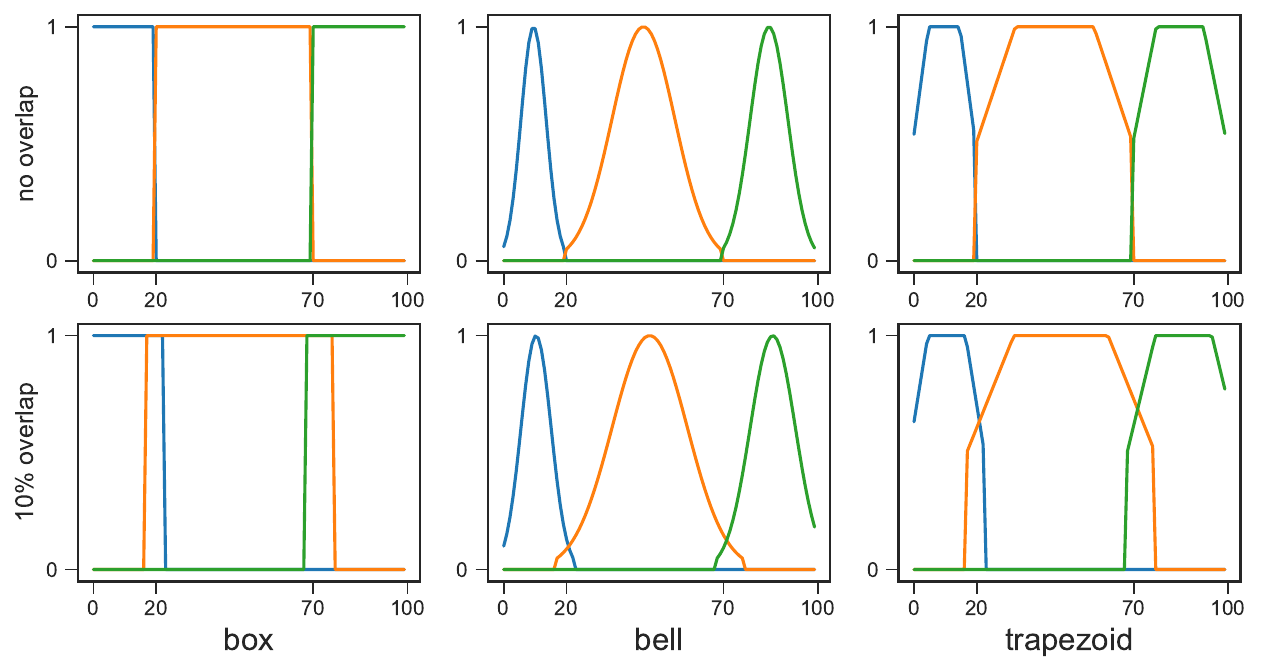}
 \caption{Examples of different masks for three consecutive action segments. The top row shows regular masks with different shapes while the bottom row shows masks generated with added $10\%$ overlap. The left, middle and right figures depict box, bell, and trapezoid shaped masks. 
 } 
 \label{fig:masks}
\end{figure}

In order to compute the mutual consistency loss, we need to generate the masks $w_m$ that act like gating functions that only allow information from the predicted temporal regions to pass through. The masks $w_m$ are functions of the predicted absolute length $\ell'_m$ and predicted starting position $p'_m$ of each segment such that
\begin{align}
 \label{eq:masks_function}
 w_m[t] &\simeq\begin{cases} 1 & p'_m\leq t\leq p'_m+\ell'_m \\ 0 & otherwise \end{cases}, \: \:
 t\in[1\dots T].
\end{align}
Examples of generated masks are shown in Figure~\ref{fig:masks}.



\subsubsection{Localization}\label{sec:loc}
In order to obtain the predicted absolute length $\ell'_m$ and the predicted starting position $p'_m$ for each segment, we first calculate the absolute length values $L'_{1:M} = (\ell'_1, \ldots, \ell'_M)$ for a video with $T$ frames such that
$\sum_{m = 1}^{M} \ell'_m = T$, \emph{i.e.}, the absolute lengths sum up to be equal to the length of the video. Having the absolute length $\ell'_m$ of each segment, we can also compute the absolute starting position $p'_m$ for each segment:
\begin{align}
 \label{eq:length_normalize}
 \ell'_m &= T \frac{e^{\ell_m}}{\sum_{k=1}^{M} e^{\ell_k}}, \:\:\:\:
 p'_1 = 0, \:\:\:\:
 p'_m = \sum_{k=1}^{m-1} \ell'_k.
\end{align}


\subsubsection{Temporal Transformation}
For generating the masks, we transform a reference template tensor $U\in[0,1]^{J}$ to $w_m\in[0,1]^T$ where $J$ is a canonical value equal to $100$.
The reference template tensor can be of any shape and we evaluate three shapes, namely \textit{box}, \textit{bell}, and \textit{trapezoid} which are depicted in Figure~\ref{fig:masks}. By adjusting the absolute lengths $\ell'_m$ and starting positions $p'_m$, it is also possible to introduce an overlap for the masks.

We transform $U$ to $w_m$ by scaling and translating it using $\ell'_m$ and $p'_m$, respectively.
Therefore, we use a 1D affine transformation matrix $\mathcal{T_\theta}$ such that
\begin{equation}
 \label{eq:affine_transformation}
 i_u[t]=\mathcal{T_\theta}(t)=[\theta_0 \; \theta_1]\begin{pmatrix} t\\1 \end{pmatrix} \quad \forall t \in [1, \dots, T]
\end{equation}
where 
$i_u[t]$ are the sampling points over $U$.
As a result of the affine transformation (\ref{eq:affine_transformation}), the element indices of the sampling points in the template $i_u[t]$ can be outside the valid range of $[1, \dots, J]$. In such cases, the indices will be ignored in the sampling process (\ref{eq:differentiable_sampling}).
The affine transformation parameters are
\begin{align}
 \label{eq:starting_position}
 \theta_0 =\frac{J}{\ell'_m}, \:\:\:\:
 \theta_1 =\frac{-Jp'_m}{\ell'_m}
\end{align}
where $\theta_0$ scales the reference template $U$ to the estimated length $\ell'_m$ and $\theta_1$ translates it to the estimated position $p'_m$.

\subsubsection{Temporal Sampling}
To perform the aforementioned temporal transformation, we should sample from $U$ using the sampling points $i_u[t]$ and produce the sampled mask $w_m$. Each index $i_u[t]$ refers to the element in $U$ where a sampling kernel must be applied to get the value at the corresponding element $w_m[t]$ in the output mask.
Similar to~\cite{STN},
we perform this operation as follows
\begin{equation}
 \label{eq:differentiable_sampling}
 w_m[t]=\sum_{j=1}^{J}U[j] \Psi(i_u[t]-j) \quad \forall t \in [1, \dots, T]
\end{equation}
where $\Psi$ is the sampling kernel. Since we use a linear kernel, it can be written as
\begin{equation}
 \label{eq:differentiable_sampling_bilinear}
 w_m[t]=\sum_{j=1}^{J}U[j] \max(0,1-|i_u[t]-j|) \quad \forall t \in [1, \dots, T].
\end{equation}

\subsubsection{Backpropagation}
To be able to backpropagate through the generated masks $w_m$, we define the gradients with respect to the sampling indices $i_u[t]$ as
\begin{equation}
 \label{eq:differentiable_sampling_bilinear_cont}
 \frac{\partial w_m[t]}{\partial i_u[t]}=
 \sum_{j=1}^{J}U_j \: \begin{cases} 0 & |i_u[t]-j| \geq 1 \\
                                    1 & j-1<i_u[t]\leq j \\
                                    -1 & j<i_u[t]<j+1 \end{cases}.
\end{equation}
Since the sampling indices $i_u[t]$ are a function of the predicted lengths $\ell'_m$, the loss gradients are backpropagated to the segment generation branch.

\subsection{Regularization}\label{sec:method:regularization}
To prevent degenerate solutions, \emph{i.e.}, solutions where the lengths of some segments are large and the length of the other segments are almost zero, we add a regularization term for the predicted relative log lengths $L$:
\begin{equation}
 \mathcal{L}_{\ell} = \sum_{m=1}^{M} \max(0,- \ell_m - w) + \max(0,\ell_m - w).
 \label{eq:length_loss}
\end{equation}
The proposed length regularizer adds a penalty if $\ell_m < -w$ or $\ell_m > w$. Note that the relative log lengths can be negative and are later converted into the absolute length as described in Section \ref{sec:loc}.  

We also use the smoothing loss introduced in \cite{MS-TCN} for the frame classification branch:
\begin{equation}
 \mathcal{L}_s = \frac{1}{TN} \sum_{t,n} \hat{\Delta}^2_{t,n},
\end{equation}
\begin{equation}
 \hat{\Delta}_{t,n} = \text{min}(\tau, |\log y_{t,n} - \log y_{t-1,n}|),
\end{equation}
with $\tau=4$ as in \cite{MS-TCN}.

\subsection{Fully Supervised and Mixed Training}\label{sec:method:full_sup}
Although our approach is designed for weakly supervised action segmentation, we can easily adapt it to a setting where all or some videos of the training set are fully annotated, \emph{i.e.}, they are annotated not by the transcript $\hat{A}$ but by the framewise labels $\hat{Y}$. 

In this case, we have two additional losses for our network shown in Figure~\ref{fig:overview}. We use the cross-entropy loss for the frame classification branch to compare the predicted class probabilities $Y$ with the framewise labels $\hat{Y}$, and we use the mean squared error loss for the segment generation branch to compare the estimated relative log lengths $L$ with the ground truth segment lengths. To this end, we convert the absolute ground-truth lengths into relative log lengths.       

Although the mutual consistency loss is only necessary in case of weakly supervised learning or in a setting where only a subset of the videos is fully annotated, we show that using the mutual consistency loss also improves the accuracy when the network is trained in a fully supervised way.     


\subsection{Inference}\label{sec:method:inference}

As discussed at the beginning of Section~\ref{sec:proposed}, we can either use the frame classification branch, which predicts the framewise class probabilities $Y$, or the segment generation branch, which predicts the segments $S$. In the later case, we start with the start symbol ${a}_{start}$ and the decoder generates new segments until the special end symbol ${a}_{end}$ is predicted as illustrated in Figure~\ref{fig:segment_gen}. The estimated relative log segment lengths are then converted into absolute lengths as described in Section \ref{sec:loc}. The two approaches for inference are denoted by MuCon-Y and MuCon-S, respectively.

However, as it is shown in Figure~\ref{fig:speed_accuracy}, we achieve the highest accuracy at a small increase in inference time if we use the predictions of both branches. We denote this approach by MuCon-full. For simplicity, we use $L$ instead of $L'$ to denote the estimated absolute lengths of the segments in this section. In order to fuse both predictions, we keep the inferred transcript $A$, but reestimate the lengths of the segments $L$ using $Y$:
\begin{equation}
 \label{eq:smoothing:def}
 L^* = \underset{\Tilde{L}}{\mathrm{argmax}}~p(\Tilde{L} | Y, A, L).
\end{equation}
Similar to~\cite{richard16,richard2018nnviterbi}, we can factorize the term $p(\Tilde{L} | Y, A, L)$ yielding
\begin{equation}
\begin{split}
 \label{eq:smoothing:rewrite}
 L^* = \underset{\Tilde{L}}{\mathrm{argmax}} \prod_{t=1}^{T} y_t[a_{\alpha(t, \Tilde{L})}] \prod_{m=1}^{M} P_{\ell_m}(\Tilde{\ell}_m)
\end{split}
\end{equation}
where $P_{\ell_m}(\Tilde{\ell}_m)$ denotes a Poisson distribution with expected mean $\ell_m$, which corresponds to the absolute segment length that has been estimated by the segment generation branch. Depending on $\Tilde{L}$, the segment number changes for a frame $t$ and it is denoted by $\alpha(t, L)$.   
\change{Although $L^*$ is obtained by dynamic programming as described in \cite{richard2018nnviterbi}, we do not need to optimize over all possible transcripts as in \cite{CDFL,richard2018nnviterbi}. While \cite{CDFL,richard2018nnviterbi} align each transcript of the training set to the test video and take the training transcript that best aligns to the test video, our approach infers the transcript $A$ directly from the test video and only aligns $A$ to the test video. MuCon-full is therefore still much faster than \cite{CDFL,richard2018nnviterbi} as shown in Figure~\ref{fig:speed_accuracy}.}

\section{Experiments}\label{sec:experiments}
We evaluate our approach for two tasks, namely action segmentation as described in Section~\ref{sec:weak_seg} and action alignment. In contrast to action segmentation, the transcripts are also given for the test videos in case of action alignment. Besides of the weakly supervised setting, we also evaluate the approach when it is trained fully supervised or in a mixed setting where some videos are fully annotated and the other videos are only weakly annotated by transcripts. Before we evaluate the approach, we discuss the evaluation protocols and further implementation details\footnote{Source code is available at: \href{https://github.com/yassersouri/MuCon}{github.com/yassersouri/MuCon}}.


\subsection{Evaluation Protocols and Datasets}\label{sec:experiments:protocol}
We evaluate our method on two popular datasets, the Breakfast dataset~\cite{breakfast} and the Hollywood extended dataset~\cite{hollywoodextended}.
The Breakfast dataset contains more than 1.7k videos of different cooking activities. The videos contain $10$ different types of breakfast activities such as \textit{prepare cereal} or \textit{prepare coffee}
which consists of $48$ different fine-grained actions. In our experiments, we follow the 4 train/test splits provided with the dataset and report the average.
The Hollywood extended dataset contains $937$ video sequences taken from Hollywood movies. The videos contain $16$ different action classes. We follow the train/test split strategy of~\cite{richard2017weakly,richard2018nnviterbi,CDFL}.

\change{The main performance metric used for the action segmentation task is the mean over frames (MoF) accuracy~\cite{MS-TCN,kuehne2016end,richard2017weakly,richard2018nnviterbi}. We also report the F1 score that has been used by fully supervised approaches~\cite{MS-TCN,mstcnpp}.
For action segmentation, we also directly evaluate the performance of the predicted transcripts, which is measured by the matching score or edit distance \cite{edit_distance}.
For action alignment, we use the intersection over detection (IoD) metric as in~\cite{hollywoodextended,richard2017weakly,richard2018nnviterbi,isba,d3tw,CDFL}. The metrics are described in the Appendix~\ref{sec:metrics}.}
If not otherwise specified, we report the average and standard deviation over 5 runs with different random seeds.  


\subsection{Implementation Details}\label{sec:experiments:details}
We train the entire network end-to-end after initializing it with Gaussian random weights. In each iteration, we use only a single video, \emph{i.e.}, the batch size is 1. We use a single group norm~\cite{groupnorm} layer with $32$ groups after the final layer of the temporal backbone $f_t$. 
The dimensionality of the shared latent video representation $Z$ is $D'=128$.
Unless otherwise stated, we apply temporal max pooling with kernel size $2$ after the convolutional layers 1, 2, 4, and 8 for all experiments on the Breakfast dataset.
\change{For the experiments on the Hollywood Extended dataset, we omit the last temporal max pooling after the convolutional layer 8 since the videos in the Hollywood extended dataset are relatively short.}
The size of the hidden states of the bidirectional LSTM encoder and the LSTM decoder in our segment generation module are set to $128$. We also employ an input embedding for the LSTM decoder of size $128$ with $0.25$ dropout.

In the weakly supervised setting, the training loss is defined as $\mathcal{L} = \mathcal{L}_{\mu} + \mathcal{L}_t + \alpha \mathcal{L}_{\ell} + \beta \mathcal{L}_s$.
We use $\alpha=\beta=0.1$ unless otherwise specified.
\change{Since most of the frames in the Hollywood extended dataset are annotated by background, which increases the possibility of degenerate solutions, we set $\alpha$ to 1 for the Hollywood extended dataset.}
We use $w=2$ in \eqref{eq:length_loss}. The initial learning rate is set to $0.01$ and is lowered by a factor of $10$ after $70$ epochs for the Breakfast dataset and after $60$ epochs for the Hollywood extended dataset. We train our network for $150$ epochs for all weakly supervised experiments. In the fully supervised and mixed supervision experiments, we train for $110$ epochs since convergence is faster with full supervision.

As input features for the Breakfast and Hollywood extended datasets, we use RGB+flow I3D~\cite{i3d} features extracted from a network that was pretrained on the Kinetics400 dataset~\cite{kinetics}. These features have been previously used for fully supervised approaches \cite{MS-TCN,mstcnpp}. A recent study found that current weakly supervised approaches perform better with IDT features than I3D features \cite{act_seg_eval2020}. We therefore report the results for the features where the corresponding methods perform best, \emph{i.e.}, I3D for \cite{MS-TCN,mstcnpp} and IDT for \cite{isba,richard2018nnviterbi,CDFL}.

\subsection{Ablation Experiments}\label{sec:experiments:ablations}

\begin{table*}[t]\centering
 \captionsetup[subfloat]{captionskip=2pt}
 \subfloat[
 \textbf{Transcript prediction accuracy}: The first four rows show the matching score of the segment generation branch for different dilation factors used for the temporal backbone. In these settings, the mutual consistency loss $\mathcal{L}_{\mu}$ is not used. The last row shows the proposed setting with the mutual consistency loss.      
 \label{tab:ablation:s2s}]{
 \small
 \begin{tabular}{m{3.5cm}|m{0.5cm}|m{1.5cm}}
 $f_t$ & $\mathcal{L}_{\mu}$ & Mat Score \\
 \Xhline{4\arrayrulewidth}
 \scriptsize{single 1-d conv} & \xmark & 0.691\mypm{0.010} \\ \hline
 \scriptsize{fixed dilation} & \xmark & 0.696\mypm{0.021} \\ \hline
 \scriptsize{increasing dilation, linear} & \xmark & 0.727\mypm{0.019} \\ \hline
 \scriptsize{increasing dilation, exponential} & \xmark & 0.729\mypm{0.015} \\
 \Xhline{4\arrayrulewidth}
 \scriptsize{increasing dilation, exponential} & \cmark & 0.774\mypm{0.006} \\
 \multicolumn{3}{c}{}
 \end{tabular}
 }
 \hspace{7mm}
 \subfloat[
 \textbf{Impact of mask generation}: \change{As shown in Figure~\ref{fig:masks}, different shapes with or without overlap can be used for the mask generation. We report the MoF accuracy, matching score, and F1 score.   }
 \label{tab:ablation:mask}]{
 \small
 \begin{tabular}{m{1.5cm}|m{1cm}|m{1cm}|m{1.5cm}|m{2.2cm}}
 Shape & Overlap & MoF & Mat Score &  F1@\{10,25,50\}\\
 \Xhline{4\arrayrulewidth}
 box & \xmark & 49.0\mypm{2.0} & 0.774\mypm{0.006} & 67.9 ~ 59.5 ~ 40.2\\ \hline
 bell & \xmark & 43.6\mypm{2.3} & 0.777\mypm{0.009} & 65.1 ~ 55.5 ~ 35.4\\ \hline
 trapezoid & \xmark & 48.7\mypm{1.8} & 0.773\mypm{0.006} & 67.3 ~ 59.0 ~ 39.4\\
 \Xhline{2\arrayrulewidth}
 box & 10\% & 48.8\mypm{1.5} & 0.787\mypm{0.006} & 68.3 ~ 59.6 ~ 39.8\\ \hline
 bell & 10\% & 43.5\mypm{0.8} & 0.779\mypm{0.007} & 64.8 ~ 55.3 ~ 34.7\\ \hline
 trapezoid & 10\% & 48.0\mypm{2.7} & 0.780\mypm{0.009} & 67.7 ~ 59.1 ~ 39.5\\
 \end{tabular}
 }

 \subfloat[
 \textbf{Impact of regularizers}: \change{The first three rows denote settings where the length regularizer $\mathcal{L}_{\ell}$ for the segment generation branch, the smoothing loss $\mathcal{L}_{s}$ for the frame classification branch, or both are omitted.}
 \label{tab:ablation:losses}]{
 \small
 \begin{tabular}{m{2.5cm}|m{1cm}|m{1.5cm}|m{2.2cm}}
 Regularizer & MoF & Mat Score & F1@\{10,25,50\} \\
 \Xhline{4\arrayrulewidth}
 none & 47.4\mypm{1.1} & 0.787\mypm{0.009} & 68.1 ~ 59.3 ~ 39.7\\ \hline
 $\mathcal{L}_{s}$ & 47.5\mypm{1.7} & 0.780\mypm{0.005} & 67.8 ~ 59.0 ~ 39.5\\ \hline
 $\mathcal{L}_{\ell}$ & 48.3\mypm{0.9} & 0.777\mypm{0.011} & 67.8 ~ 59.0 ~ 40.3\\
 \Xhline{4\arrayrulewidth}
 $\mathcal{L}_{\ell}$ + $\mathcal{L}_{s}$ & 49.0\mypm{2.0} & 0.774\mypm{0.006} &  67.9 ~ 59.5 ~ 40.2\\
 \end{tabular}
 }
 \hspace{7mm} 
 \subfloat[
 \change{
 \textbf{Impact of fusing both branches}: While MuCon-Y and MuCon-S use either the frame classification branch or the segment generation branch for inference, MuCon-full fuses both branches for inference.
 }
 \label{tab:ablation:variants}]{
 \small
 \begin{tabular}{m{3.0cm}|m{1cm}|m{2.2cm}}
 Inference variant & MoF &   F1@\{10,25,50\}\\
 \Xhline{4\arrayrulewidth}
 MuCon-Y & 44.7\mypm{1.4} &  28.1 ~ 22.6 ~ 13.3\\ \hline
 MuCon-S & 43.6\mypm{1.2} &  65.6 ~ 57.2 ~ 33.8\\
 \Xhline{4\arrayrulewidth}
 MuCon-full & 49.0\mypm{2.0} &   67.9 ~ 59.5 ~ 40.2\\
 \multicolumn{3}{c}{}
 \end{tabular}
 }
 
\subfloat[
 \textbf{Impact of teacher forcing}: \change{The first row denotes a setting where teacher forcing is not used for training. The second row denotes a setting where teacher forcing is used only for the first 70 epochs. The last row denotes a setting where teacher forcing is used for all epochs.}
 \label{tab:ablation:tf}]{
 \small
 \begin{tabular}{m{3cm}|m{1cm}|m{1.5cm}|m{2.2cm}}
 Teacher forcing & MoF & Mat Score & F1@\{10,25,50\} \\
 \Xhline{4\arrayrulewidth}
 None & 48.7\mypm{2.9} & 0.779\mypm{0.009} & 68.2 ~ 59.5 ~ 40.3\\ \hline
 70 epochs & 48.9\mypm{2.1} & 0.783\mypm{0.009} & 68.2 ~ 59.5 ~ 39.8\\ \Xhline{4\arrayrulewidth}
 All epochs & 49.0\mypm{2.0} & 0.774\mypm{0.006} &  67.9 ~ 59.5 ~ 40.2\\
 \end{tabular}
 }

 \vspace{5mm}
 \caption{Ablation experiments on split 1 of the Breakfast dataset. We report the average and standard deviation over 5 runs.}
 \label{tab:ablation}
 \end{table*}

In this section, we quantitatively examine different components in our method. For each metric, We report the average and standard deviation over 5 runs on split 1 of the Breakfast dataset \cite{breakfast}.

\subsubsection{Transcript Prediction}
We first analyze how well the proposed approach predicts the transcripts, which is measured by the matching score. If we only use the segment generation branch, the network achieves a matching score of $0.729$ as shown in row 4 of Table~\ref{tab:ablation:s2s}. If we add the frame classification branch and the mutual consistency loss $\mathcal{L}_{\mu}$, the matching score increases to $0.774$. This shows that the mutual consistency loss also improves the transcript prediction.    

In the first four rows of Table~\ref{tab:ablation:s2s}, we furthermore evaluate the impact of the dilation factors for the temporal backbone $f_t$. If we replace the temporal backbone by a single 1-d convolution, the matching score decreases from $0.729$ to $0.691$. If we use as in the proposed network 11 layers with 1-d convolutions but with fixed dilation factor of $1$, the matching score does not significantly increase since it does not substantially increase the receptive field. Only when we linearly or exponentially increase the dilation factors for each layer, we observe an improvement of the matching score.


\subsubsection{Mask Generation Settings}

As described in Section~\ref{sec:mask}, different shapes of mask templates can be used for the mutual consistency loss. We evaluated three shapes, namely \textit{box}, \textit{bell}, and \textit{trapezoid}, and added an optional overlap of 10\% as depicted in Figure~\ref{fig:masks}. The results reported in Table~\ref{tab:ablation:mask} show that the shape of the mask has little impact on the quality of the predicted transcript, but that the bell shape has the worst performance. In general, having hard boundaries for the masks without any overlap performs best. For all other experiments, we therefore use the \textit{box} shape without any overlap.




\subsubsection{Effect of Regularizers}

As described in Section~\ref{sec:method:regularization}, we also use two additional regularizers during training, namely the length loss $\mathcal{L}_{\ell}$ and the smoothing loss $\mathcal{L}_s$. \change{We evaluate the impact of each term in Table~\ref{tab:ablation:losses} by removing $\mathcal{L}_{\ell}$, $\mathcal{L}_s$, or both. Without any regularizer, the MoF accuracy decreases from $49.0$ to $47.4$ and F1@50 from $40.2$ to $39.7$. While adding only the smoothing loss does not result in an improvement, the length regularizer and the combination of both improves MoF.         
} 

\subsubsection{Different Variants for Inference}

As described in Section~\ref{sec:method:inference}, we have three options for inference. We can use the frame classification branch, the segment generation branch, or both. The three approaches are denoted by MuCon-Y, MuCon-S, and MuCon-full, respectively. \change{The results in Table~\ref{tab:ablation:variants} show that fusing the two branches improves the MoF accuracy compared to MuCon-S by more than $5\%$ and F1@50 by more than $6\%$ at a small increase in inference time as shown in Figure~\ref{fig:speed_accuracy}. Since MuCon-Y predicts only frame-wise labels and not segments, the F1 scores are very low for MuCon-Y.}   

\subsubsection{\change{Effect of Teacher Forcing}}
\change{
As described in Sections~\ref{sec:method:fs} and \ref{sec:method:mucon}, we use the ground truth transcript while predicting the transcript and calculating the MuCon loss in order to stabilize the training at the beginning. In Table~\ref{tab:ablation:tf}, we evaluate the impact of teacher forcing where we also include a setting that uses teacher forcing only for the first 70 epochs of training. The results show that teacher forcing does not impact the average accuracy but it reduces the standard deviation, which is an indicator of a stable training procedure.  
}

\subsection{Statistical Analysis}\label{sec:experiments:stats}

\begin{table*}[t]\centering
 \small
 \begin{tabular}{l|c|c|c|c}
 \multicolumn{3}{c}{}\\
 Approach & \multicolumn{1}{p{2cm}|}{\centering Avg MoF\mypm{Std}}
 & \multicolumn{1}{p{3cm}|}{\centering Avg Mat Score\mypm{Std}} & \multicolumn{1}{p{2.6cm}|}{\centering Training (hours)} & \multicolumn{1}{p{2.75cm}}{\centering Inference (seconds)}\\
 \Xhline{4\arrayrulewidth}
 ISBA \cite{isba} & 36.4\mypm{1.0} & - & 12.75 & 0.01 \\
 NNV \cite{richard2018nnviterbi} & 39.7\mypm{2.4} & 0.686\mypm{0.009} & 11.23 & 56.25 \\
 CDFL \cite{CDFL} & 48.1\mypm{2.5} & 0.712\mypm{0.009} & 66.73 & 62.37\\
 \hline
 MuCon-Y & 44.2\mypm{1.4} & - & 4.57 & 0.02\\
 MuCon-S & 43.9\mypm{1.4} & \textbf{0.785}\mypm{0.008} & 4.57 & 0.04\\
 MuCon-full & \textbf{48.5}\mypm{1.8} & \textbf{0.785}\mypm{0.008} & 4.57 & 3.03\\
 \end{tabular}
\caption{Comparison of accuracy, training, and inference time. We report the average and standard deviation of MoF and matching score for 5 runs on the entire Breakfast dataset~\cite{breakfast}. For ISBA and MuCon-Y, the matching score is not calculated as they predict frame-wise labels and not the transcript. Training and inference time is measured as wall clock time. The training time is measured for training on split 1 of the Breakfast dataset~\cite{breakfast}. The inference time is measured as the average inference time for a video from the test set of split 1.
}
 \label{tab:further-exp-stat}\label{tab:further-exp:reprod}\label{tab:further-exp:speed}
 \end{table*}

As the standard deviations in Table \ref{tab:ablation} show, the results vary for different random seeds. In order to provide a thorough comparison with other weakly supervised action segmentation approaches, we use a statistical test to measure if the differences in accuracy are significant or not. To this end, we performed five training/inference iterations on the entire Breakfast dataset. Besides of our approach (MuCon), we included CDFL~\cite{CDFL}, NNV~\cite{richard2018nnviterbi}, and ISBA~\cite{isba} for which the official open-source implementation is available. We use the hyperparameters as they were provided by the authors for this dataset. 
We use the Friedman statistical test \cite{friedmantest} with the dataset split as the blocking factor.
We selected the Friedman test as it does not make any restrictive assumption on the underlying distribution of the data. 

While we report the average and standard deviation of the MoF accuracy and matching score in Table~\ref{tab:further-exp:reprod}, the results of the Friedman test for the MoF accuracy and the matching score are shown in Tables~\ref{tab:further-exp:stat-test:mof} and \ref{tab:further-exp:stat-test-matscore}, respectively. We observe that MuCon and CDFL are both significantly better than ISBA and NNV. While the average MoF accuracy and the average matching score of MuCon are higher than the ones of CDFL, the difference is only statistically significant for the matching score. This shows that MuCon is on par with CDFL in terms of MoF accuracy, but significantly better in terms of matching score.

\begin{table*}[t]\centering
 \captionsetup[subfloat]{captionskip=2pt}
 \subfloat[
 Friedman statistical test for MoF.
 \label{tab:further-exp:stat-test:mof}]{
 \tiny
 \begin{tabular}{c?c|c|c|c}
 p-value&
 ISBA (36.4) &
 NNV (39.7) & 
 CDFL (48.1) & 
 MuCon (48.5)\\
 \Xhline{4\arrayrulewidth}
 ISBA (36.4) & & \cmark (0.001) & \cmark ($1.77$\smallmul{$10^{-7}$}) & \cmark ($1.77$\smallmul{$10^{-7}$})\\
 \hline
 NNV (39.7) &\cmark (0.001) & & \cmark ($3.09$\smallmul{$10^{-7}$}) & \cmark ($5.35$\smallmul{$10^{-7}$})\\
 \hline
 CDFL (48.1) &\cmark ($1.77$\smallmul{$10^{-7}$}) & \cmark ($3.09$\smallmul{$10^{-7}$}) & & \xmark (0.465) \\
 \hline
 MuCon (48.5) &\cmark ($1.77$\smallmul{$10^{-7}$}) & \cmark ($5.35$\smallmul{$10^{-7}$}) & \xmark (0.465) & \\
 \end{tabular}
 }
 \hspace{5mm}
 \subfloat[
 Friedman statistical test for matching score.
 \label{tab:further-exp:stat-test-matscore}]{
 \tiny
 \begin{tabular}{c?c|c|c}
 p-value&
 NNV (0.686) & 
 CDFL (0.712) & 
 MuCon (0.785)\\
 \Xhline{4\arrayrulewidth}
 NNV (0.686) & & \cmark ($1.96$\smallmul{$10^{-6}$}) & \cmark ($1.77$\smallmul{$10^{-7}$})\\
 \hline
 CDFL (0.712) & \cmark ($1.96$\smallmul{$10^{-6}$}) & & \cmark ($1.77$\smallmul{$10^{-7}$})\\
 \hline
 MuCon (0.785) & \cmark ($1.77$\smallmul{$10^{-7}$}) & \cmark ($1.77$\smallmul{$10^{-7}$}) & \\
 \multicolumn{4}{c}{}
 \end{tabular}
 }
 \vspace{5mm}
 \caption{Friedman statistical test results.
 The top row and the leftmost column indicate different approaches and their MoF accuracy and matching score from Table~\ref{tab:further-exp:reprod}.
 Note that the test is symmetrical.
 If the p-value is smaller than $0.05$ the difference is considered as significant (indicated with \cmark); otherwise the difference is not considered as significant (indicated with \xmark).}
 \label{tab:further-exp:stat-test}
\end{table*}

\subsection{Training and Inference Time}\label{sec:experiments:speed}
Besides of the accuracy, we also compare the training and inference time of our approach with CDFL~\cite{CDFL}, NNV~\cite{richard2018nnviterbi}, and ISBA~\cite{isba}. For a fair comparison, we always used the same hardware. For training, we used a machine with an Nvidia GeForce GTX 1080Ti GPU, 188 GB of RAM, and an Intel(R) Xeon(R) Gold 5120 (2.20GHz) CPU. For testing, we used a machine with an Nvidia GeForce GTX Titan X GPU, 32 GB of RAM, and an Intel(R) Core(TM) i7-4930K (3.40GHz) CPU. We only calculate the wall time for training and deactivated for all methods any unnecessary operations like saving intermediate results. \change{Since we used pre-computed features for all experiments, the time measurement includes the time to load the features but not the time to compute the features. In average, computing the features takes 92 seconds  per video for the Breakfast dataset where 72.5 seconds are spent for calculating the optical flow.}  


The results are reported in Table~\ref{tab:further-exp:speed}.
We observe that our approach is 14 times faster to train and 20 times faster during inference compared to the state-of-the-art approach CDFL~\cite{CDFL}.
As mentioned before, our approach does not perform any Viterbi decoding during training, which makes it faster during training.
Also and most importantly, our approach only performs one Viterbi decoding step for the estimated transcript during inference as compared to CDFL and NNV which need to optimize over all possible transcripts. This shows that our approach offers by far the best trade-off between accuracy and runtime as it is also illustrated in Figure~\ref{fig:speed_accuracy}.


\begin{figure}[t]
 \centering
 \includegraphics[width=0.48\textwidth]{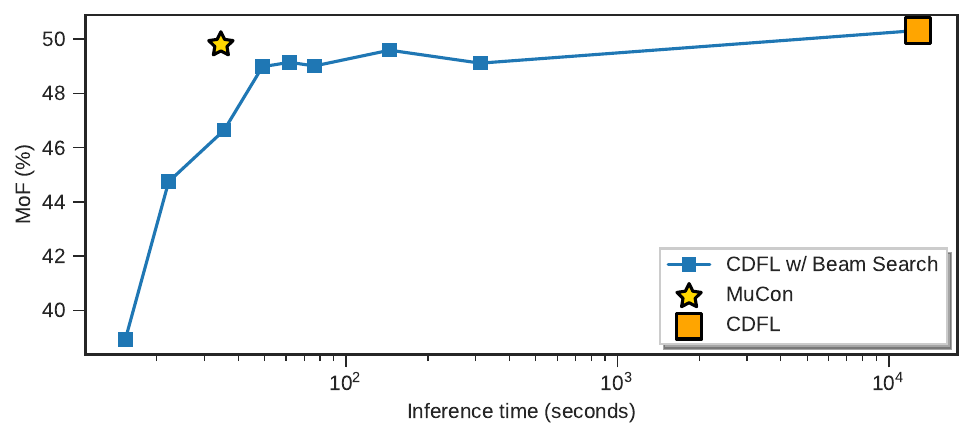}
 \vspace{-2mm}
 \caption{
    \change{
        Using beam search to reduce the inference time of CDFL. The blue squares show the performance of CDFL with different values for the beam size. The x-axis is the total inference time for split 1 of the Breakfast dataset in log scale.
    }
 }
 \label{fig:beam_search}
\end{figure}

\change{In order to avoid the full alignment of all transcripts of the training set to a test video, beam search can be used. Beam search limits the maximum number of hypotheses and removes hypotheses with low probability at an early stage. The smaller the beam size, i.e., the number of hypotheses, is, the faster is the inference. This, however, comes at the cost of reducing the accuracy. In order to evaluate how much runtime reduction can be achieved by beam search, we evaluate CDFL with beam search. Figure~\ref{fig:beam_search} shows the MoF accuracy and inference time for different beam sizes. In contrast to Figure~\ref{fig:speed_accuracy}, we report the accuracy for only one run and only split 1 of the Breakfast dataset. We furthermore plot the inference time of the entire test set in log scale for better visualization. We observe that beam search reduces the inference time of CDFL by multiple orders of magnitude but also its accuracy. When we compare a setting where CDFL with beam search is as fast as MuCon, we observe that MuCon achieves a much higher accuracy.  
}

\subsection{Further Comparisons}\label{sec:experiments:comparison}

\begin{table*}[t]\centering
 \captionsetup[subfloat]{captionskip=2pt}
 \subfloat[
 Comparison on the Breakfast dataset.  
 \label{tab:comparison:breakfast}]{
 \small
 \begin{tabular}{m{2.5cm}|m{1.95cm}|m{1.75cm}}
 Approach & MoF - Action Segmentation & IoD - Action Alignment\\
 \Xhline{4\arrayrulewidth}
 ECTC \cite{ectc} & 27.7 & 45.0 \\
 HMM\slash RNN \cite{richard2017weakly} & 33.3 & 47.3\\
 ISBA \cite{isba} & 38.4 & 52.3\\
 NNV \cite{richard2018nnviterbi} & 43.0 & -\\
 D3TW \cite{d3tw} & 45.7 & 56.3\\
 CDFL \cite{CDFL} & \textbf{50.2} & \underline{63.9}\\
 \hline
 MuCon$\ast$ & \underline{48.5} & \textbf{66.2}\\
 \end{tabular}
 }
 \hspace{3mm}
 \subfloat[
 Comparison on the Hollywood extended dataset. 
 \label{tab:comparison:hollywoodextended}]{
 \small
 \begin{tabular}{m{2.5cm}|m{2.5cm}|m{1.75cm}}
 Approach & MoF-BG - Action Segmentation & IoD - Action Alignment\\
 \Xhline{4\arrayrulewidth}
 ECTC \cite{ectc} & - & 41.0 \\
 HMM\slash RNN \cite{richard2017weakly} & - & 46.3\\
 ISBA \cite{isba} & 34.5 & 39.6\\
 NNV \cite{richard2018nnviterbi} & - & 48.7\\
 D3TW \cite{d3tw} & 33.6 & 50.9\\
 CDFL \cite{CDFL} & \underline{40.6} & \textbf{52.9}\\
 \hline
 MuCon$\ast$ & \textbf{41.6} & \underline{52.3}\\
 \end{tabular}
 }
 \vspace{1mm}
 \caption{Comparison to the state-of-the-art. The highest score is indicated by a bold font and the second highest is underlined. $\ast$ indicates that the reported metric is averaged over 5 runs.}
 \label{tab:comparison}
 \end{table*}

Although a statistical test is the golden standard for comparing different methods, this is only possible if the results of multiple runs are available. We therefore also compare our method with state-of-the-art methods based on the reported numbers on the Breakfast dataset~\cite{breakfast} in Table~\ref{tab:comparison:breakfast} and on the Hollywood extended dataset~\cite{hollywoodextended} in Table~\ref{tab:comparison:hollywoodextended} for weakly supervised action segmentation and alignment. Our approach outperforms all other methods except of CDFL for both tasks and datasets. As already discussed in Sections~\ref{sec:experiments:stats} and \ref{sec:experiments:speed}, our approach achieves either an accuracy that is comparable to CDFL or significantly better. Furthermore, our approach is 20 times faster during inference and provides thus a much better trade-off between accuracy and runtime.


\subsubsection{Fully Supervised}\label{sec:experiments:full_sup}
As mentioned in Section~\ref{sec:method:full_sup}, we can apply our approach to the fully supervised setting as well. For comparison, we use the same metrics as in \cite{MS-TCN,mstcnpp} namely MoF (termed accuracy in \cite{MS-TCN}), Edit which measures predicted transcript similarity (similar to the matching score), and F1 score at different overlaps.

We first evaluate whether the mutual consistency loss $\mathcal{L}_{\mu}$ also improves the accuracy in a fully supervised setting. For this ablation experiment, we use split 1 of the Breakfast dataset. As it is shown in Table~\ref{tab:full_sup:ablation}, the proposed mutual consistency loss improves the accuracy for all metrics. This shows that the mutual consistency loss is not only useful for weakly supervised learning, but also for fully supervised learning.  

We furthermore compare our approach to other fully supervised action segmentation approaches in Table~\ref{tab:full_sup}. Although our approach was designed for weakly supervised learning, it outperforms the state-of-the-art for most metrics. Only for the MoF accuracy, \cite{MS-TCN,mstcnpp} perform better, but these networks use multiple stages and thus more layers.      

\begin{table}
 \centering
 \resizebox{\linewidth}{!}{
 \begin{tabular}{lccccc}
 \toprule
 \textbf{Breakfast} & \multicolumn{3}{c}{F1@\{10,25,50\}} & Edit & MoF \\ 
 \midrule
 MuCon w/o $\mathcal{L}_{\mu}$ & 71.6\mypm{0.9} & 64.8\mypm{0.8} & 49.1\mypm{1.1} & 75.6\mypm{0.9} & 63.7\mypm{0.8} \\ 
 MuCon & \textbf{73.1}\mypm{0.4} & \textbf{66.3}\mypm{0.6} & \textbf{50.7}\mypm{1.2} & \textbf{76.4}\mypm{0.6} & \textbf{64.1}\mypm{1.1} \\ 
 \bottomrule
 \end{tabular}
 }
 \vspace{1mm}
 \caption{Effect of the MuCon loss on the accuracy in the fully supervised setting. The results are reported for split 1 of the Breakfast dataset.}
 \label{tab:full_sup:ablation}
\end{table}

\begin{table}
 \centering
 \resizebox{\linewidth}{!}{
 \begin{tabular}{lccccc}
 \toprule
 \textbf{Breakfast} & \multicolumn{3}{c}{F1@\{10,25,50\}} & Edit & MoF 
 \\ \midrule
 ED-TCN~\cite{lea2017temporal}* & - & - & - & - & 43.3 \\
 HTK~\cite{hildecviu} & - & - & - & - & 50.7 \\
 TCFPN~\cite{isba} & - & - & - & - & 52.0 \\
 HTK(64)~\cite{kuehne2016end} & - & - & - & - & 56.3 \\ 
 GRU~\cite{richard2017weakly}* & - & - & - & - & 60.6 \\ 
 GRU+length prior~\cite{kuehne2018hybrid} & - & - & - & - & 61.3 \\ 
 MS-TCN~\cite{MS-TCN} & 52.6 & 48.1 & 37.9 & 61.7 & 66.3 \\ 
 MS-TCN++~\cite{mstcnpp} & 64.1 & 58.6 &45.9 & 65.6 & \textbf{67.6} \\
 \midrule
 MuCon & \textbf{73.2}\mypm{0.4} & \textbf{66.1}\mypm{0.5} & \textbf{48.4}\mypm{0.6} & \textbf{76.3}\mypm{0.5} & 62.8\mypm{1.0} \\ 
 \bottomrule
 \end{tabular}
 }
 \vspace{1mm}
 \caption{Comparison with the state-of-the-art for fully supervised action segmentation on the Breakfast dataset. (* obtained from~\cite{isba}).}
 \label{tab:full_sup}
\end{table}

\subsubsection{Mixed Supervision}\label{sec:experiments:mixed_sup}

\begin{figure}
 \centering
 \includegraphics[width=0.48\textwidth]{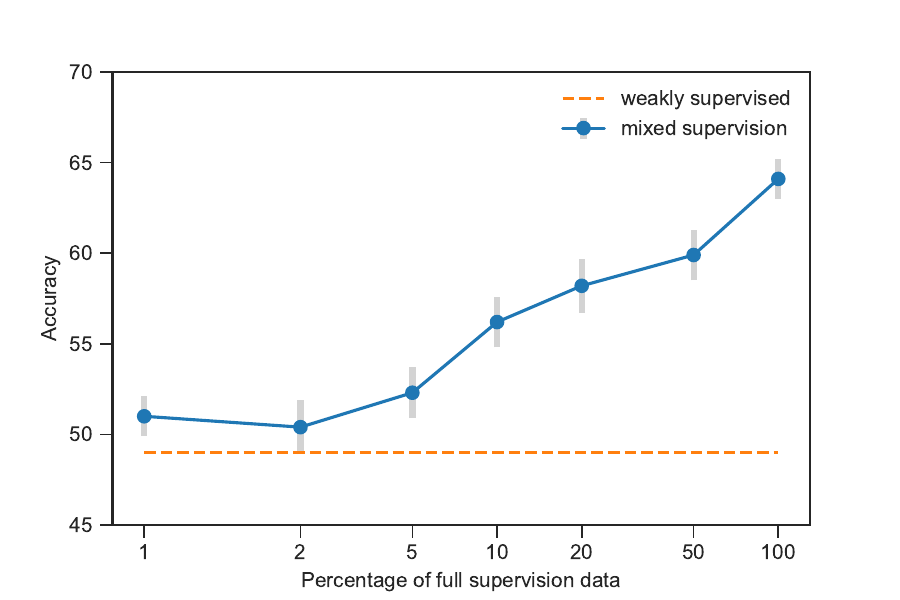}
 \caption{MoF accuracy (\%) for training with mixed supervision. The x-axis denotes the percentage of videos that are fully annotated. The average accuracy for 5 runs on split 1 of the Breakfast dataset is reported.}
 \label{fig:mixed}
\end{figure}

Since our network can be trained in a weakly as well in a fully supervised setting, we can train the network also in a mixed setting where a small percentage of the videos is fully annotated and the remaining videos are only weakly annotated by transcripts. As before, we report the average and standard deviation over 5 runs where we randomly sample the videos with frame-wise annotations for each run. The results for a varying percentage of fully annotated videos are reported in Table~\ref{tab:mixed} and visualized in Figure~\ref{fig:mixed}. In case of 0\%, the setting corresponds to weakly supervised learning and 100\% corresponds to fully supervised learning. The difference in accuracy between the weakly and fully supervised cases is about 15\%. While there is no significant improvement if only 1-2\% of the videos are fully annotated, the accuracy increases when at least 5\% of the videos are fully annotated. Having 10\% of the videos fully annotated, the accuracy gap between the mixed setting and the fully supervised setting is already reduced by about 50\%.

\begin{table*}
 \centering
 \begin{tabular}{c?c|c|c|c|c|c|c|c}
 Percentage of fully supervised data $p\%$ & 0\% & 1\% & 2\% & 5\% & 10\% & 20\% & 50\% & 100\%\\
 \Xhline{4\arrayrulewidth}
 Accuracy \% & 49.0\mypm{2.0} & 51.0\mypm{1.1} & 50.4\mypm{1.5} & 52.3\mypm{1.4} & 56.2\mypm{1.4} & 58.2\mypm{1.5} & 59.9\mypm{1.4} & 64.1\mypm{1.1}
 \end{tabular}
 \vspace{5mm}
 \caption{Training with mixed supervision. The average MoF accuracy and standard deviation for 5 runs on split 1 of the Breakfast dataset is reported.}
 \label{tab:mixed}
\end{table*}

\subsection{\change{Qualitative Evaluation}}\label{sec:experiments:qual}

\begin{figure}
    \centering
    \subfloat[
        \mybox{c1}~\textit{background} -
        \mybox{c2}~\textit{spoon powder} -
        \mybox{c3}~\textit{pour~milk} -
        \mybox{c4}~\textit{stir~milk} -
        \mybox{c5}~\textit{take~cup}
    ]{
        \includegraphics[width=0.5\textwidth]{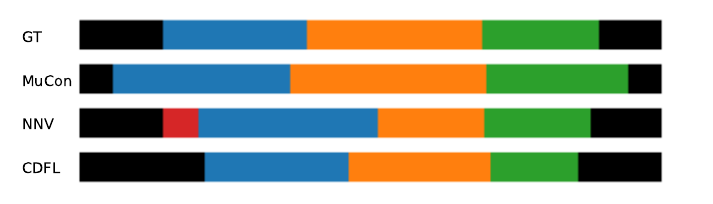}
        \label{fig:qual:1}
    }\\
    \subfloat[
        \mybox{c1}~\textit{background} -
        \mybox{c2}~\textit{pour~cereals} -
        \mybox{c3}~\textit{pour~milk} -
        \mybox{c4}~\textit{take~bowl} -
        \mybox{c5}~\textit{stir~cereals}
    ]{
        \includegraphics[width=0.5\textwidth]{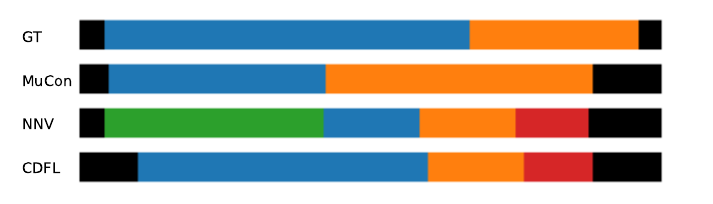}
        \label{fig:qual:3}
    }\\
    \subfloat[
        \mybox{c1}~\textit{background} -
        \mybox{c2}~\textit{pour~coffee} -
        \mybox{c3}~\textit{pour~milk} -
        \mybox{c4}~\textit{take~cup} -
        \mybox{c5}~\textit{add~teabag} -
        \mybox{c6}~\textit{pour~water} -
        \mybox{c7}~\textit{spoon~sugar}
    ]{
        \includegraphics[width=0.5\textwidth]{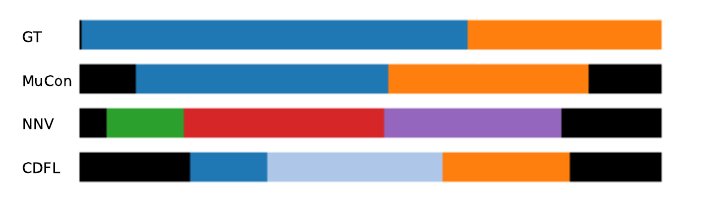}
        \label{fig:qual:4}
    }\\
    \subfloat[
        \mybox{c1}~\textit{background} -
        \mybox{c2}~\textit{take~plate} -
        \mybox{c3}~\textit{take~knife} -
        \mybox{c4}~\textit{cut~orange} -
        \mybox{c5}~\textit{squeeze~orange} -
        \mybox{c6}~\textit{take~glass} -
        \mybox{c7}~\textit{pour~juice}
    ]{
        \includegraphics[width=0.5\textwidth]{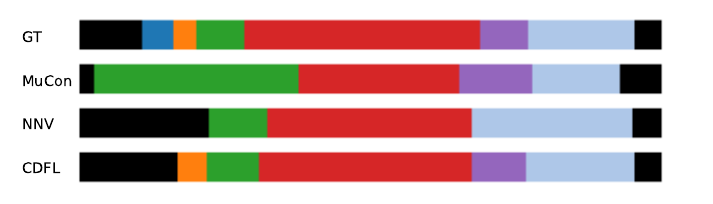}
        \label{fig:qual:5}
    }
    \caption{
    \change{Qualitative examples for weakly supervised action segmentation on the Breakfast dataset. Each figure visualizes a different video from the test set of split 1. We compare the results from MuCon, NNV, and CDFL with the ground truth (GT). Each row shows the result for the entire duration of a video and the colored segments show when the actions occur in the video. }
    }
    \label{fig:qual}
\end{figure}

\change{
We provide some qualitative results for four different test videos of the Breakfast dataset in Figure~\ref{fig:qual}. Figure~\ref{fig:qual:1} shows a video where NNV, CDFL, and MuCon perform well. Only NNV hallucinates the action \emph{take cup}, which is not present in the video. The hallucination of actions occurs due to the alignment of the transcripts of the training set to the test video. In this example, there is a very high uncertainty among the class probabilities estimated by NNV at the beginning of the video. Since the probabilities for \emph{take cup} are low but slightly higher than for \emph{spoon powder} or \emph{background}, the transcript \emph{take cup - spoon powder - pour milk - stir milk} achieves a higher alignment score than the correct transcript \emph{spoon powder - pour milk - stir milk}. Figures~\ref{fig:qual:3} and \ref{fig:qual:4} show examples where also CDFL hallucinates actions that are plausible based on the transcripts but that do not occur in the video. In contrast, MuCon does not suffer from hallucinating actions since it infers the transcript directly from the test video and it does not search the training transcript that best aligns to the test video. Figure~\ref{fig:qual:5} visualizes a failure case where all approaches fail to infer the \emph{take plate} action at the beginning of the video. In this example, CDFL provides a better estimate than MuCon.
}

\begin{figure}
    \centering
    \subfloat[
        \mybox{c1}~\textit{background} -
        \mybox{c2}~\textit{add~teabag} -
        \mybox{c3}~\textit{pour~water}
    ]{
        \includegraphics[width=0.5\textwidth]{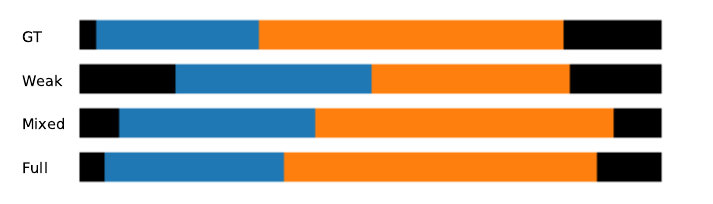}
        \label{fig:qual:sup:1}
    }\\
    \subfloat[
        \mybox{c1}~\textit{background} -
        \mybox{c2}~\textit{pour~oil} -
        \mybox{c3}~\textit{crack~egg} -
        \mybox{c4}~\textit{fry~egg} -
        \mybox{c5}~\textit{add~salt~and~pepper} -
        \mybox{c6}~\textit{put~egg~to~plate}
    ]{
        \includegraphics[width=0.5\textwidth]{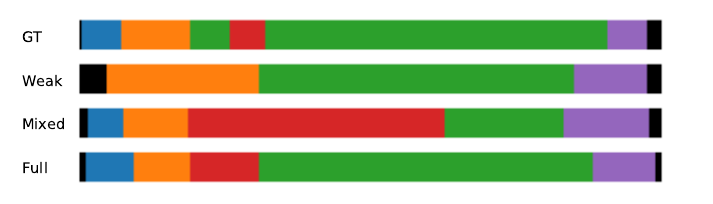}
        \label{fig:qual:sup:2}
    }
    \caption{
        \change{
            Qualitative examples for different levels of supervision on the Breakfast dataset. Each figure visualizes a different video from the test set of split 1. \textit{GT} visualizes the ground truth segmentation and \textit{Weak}, \textit{Mixed}, and \textit{Full} visualize the output of MuCon trained with weak, mixed (10\%), or full supervision, respectively.
        }
    }
    \label{fig:qual:sup}
\end{figure}

\change{
Figure~\ref{fig:qual:sup} shows qualitative examples for comparing the different types of supervision, namely weak, mixed (10\%), and full supervision.
Figure~\ref{fig:qual:sup:1} shows the result for a test video where MuCon is able to infer the actions correctly using weak, mixed, or full supervision. However, the action boundaries are more accurately estimated if MuCon is trained with more supervision.
Figure~\ref{fig:qual:sup:2} shows a very difficult video where even the fully supervised approach makes a mistake.  
}

\section{Conclusion}
We proposed a novel approach for weakly supervised action segmentation from transcripts. It consists of a two-branch network that predicts two representations for action segmentation. To train the network, we introduced a new mutual consistency loss (MuCon) that enforces these two representations to be consistent during training. Using MuCon and a transcript prediction loss, we can train our network end-to-end without the need for additional steps. We show that the proposed network with the mutual consistency loss achieves an accuracy that is either on par or better than the state-of-the-art. At the same time, it is much faster during inference and offers the best trade-off between accuracy and inference time. Furthermore, we have shown that the mutual consistency loss increases the accuracy even for fully supervised learning and that the network can be applied to a mixed setting where a few videos are fully annotated and the other videos are only weakly annotated.



%

\appendices

\section{\change{Metrics}}\label{sec:metrics}

\subsection{\change{Mean over Frames Accuracy (MoF)}}
\change{
The mean over frames accuracy~\cite{MS-TCN,kuehne2016end,richard2017weakly,richard2018nnviterbi} is defined as the number of frames with correctly predicted labels divided by the total number of frames.
Given a set of predictions $Y^{(j)}$ and ground truth labels $\hat{Y}^{(j)}$, where $j = 1, \dots, V$ and $V$ is the number of videos in the test set, the MoF is defined as
\begin{equation}
 \text{MoF} = \frac{\displaystyle\sum_{j=1}^{V} \displaystyle\sum_{t=1}^{T^{(j)}} \mathbb{I}(Y_t^{(j)} = \hat{Y}_t^{(j)})}{\displaystyle\sum_{j=1}^{V} T^{(j)}}
\end{equation}
where $T^{(j)}$ is the length of the $j$th video and $\mathbb{I}$ is the indicator function.
}

\change{
MoF-BG is similar to MoF, but it only considers the ground truth frames which are not annotated as background (BG).
}

\subsection{\change{Matching Score}}
\change{
The matching score \cite{edit_distance} measures the similarity of the predicted transcript compared to the ground truth transcript:
\begin{equation}
 \text{Matching Score} = 2 \times \frac{\text{number of matches}}{\vert A \vert + \vert \hat{A} \vert }
\end{equation}
where $\vert A \vert $ is the length of the predicted transcript and $\vert \hat{A} \vert$ is the length of the ground truth transcript.
}

\subsection{\change{Intersection Over Detection (IoD)}}

\change{
IoD is only defined for the alignment task since the transcript and number of segments are known for this task. This provides a one to one matching between the predicted and ground truth segments. The IoD is then computed by the average intersection of a ground truth segment with an associated predicted segment divided by the length of the predicted segment, i.e., $\vert S \cap \hat{S} \vert / \vert {S} \vert$. We use the same definition and code as \cite{isba,CDFL}.
}

\subsection{\change{Edit and F1 Score}}

\change{
Edit and F1 scores are calculated using the same definition and code as \cite{MS-TCN,mstcnpp}.
}


\ifCLASSOPTIONcompsoc
  \section*{Acknowledgments}
\else
  \section*{Acknowledgment}
\fi

The work has been partially funded by the Deutsche Forschungsgemeinschaft (DFG, German Research Foundation) - GA 1927/4-2 (FOR 2535 Anticipating Human Behavior) and the ERC Starting Grant ARCA (677650).

\ifCLASSOPTIONcaptionsoff
  \newpage
\fi



%

\bibliographystyle{IEEEtran}
\bibliography{references}

%

\begin{IEEEbiography}[{\includegraphics[width=1in,height=1.25in,clip,keepaspectratio]{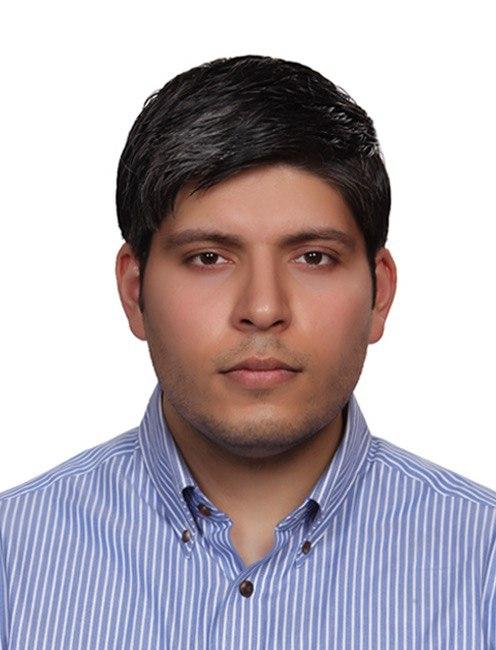}}]{Yaser Souri}
received his Bachelor degree in Computer Engineering from Iran University of Science and Technology in 2011 and his Master degree in Artificial Intelligence from Sharif University of Technology in 2015. Since 2017, he is a PhD student at the University of Bonn. His research interests include action segmentation and video understanding.
\end{IEEEbiography}

\begin{IEEEbiography}[{\includegraphics[width=1in,height=1.25in,clip,keepaspectratio]{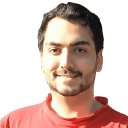}}]{Mohsen Fayyaz}
obtained his Bachelor degree in Software Engineering from Semnan University in 2014 and his Master degree in Artificial Intelligence from Malek-Ashtar University of Technology in 2016. Since 2017, he is a Ph.D. candidate at the Computer Vision Group of the University of Bonn. His research interests include video understanding, action segmentation, and action recognition.
\end{IEEEbiography}

\begin{IEEEbiography}[{\includegraphics[width=1in,height=1.25in,clip,keepaspectratio]{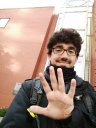}}]{Luca Minciullo} obtained his Bachelor and Master degrees from the ``La Sapienza'' University of Rome, in 2012 and 2014, respectively. In 2014, he started a PhD at the University of Manchester in computer vision and medical imaging under the supervision of Prof. T.F. Cootes. Since 2018, he is an Engineer at Toyota Motor Europe, developing perception technologies for indoor robotics.
\end{IEEEbiography}

\begin{IEEEbiography}[{\includegraphics[width=1in,height=1.25in,clip,keepaspectratio]{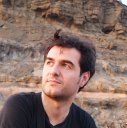}}]{Gianpiero Francesca} obtained his Bachelor and Master degrees from the ``Sannio'' University in Benevento, Italy. In 2011 he started a PhD at the ULB University in Brussels, focusing on swarm robotics and automatic design of control software, under the supervision of Prof. M. Birattari. Since 2015, he has been working for Toyota Motor Europe. Today he is a Senior Engineer there, developing human activity recognition technology.
\end{IEEEbiography}

\begin{IEEEbiography}[{\includegraphics[width=1in,height=1.25in,clip,keepaspectratio]{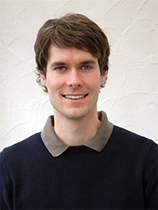}}]{Juergen Gall}
obtained his B.Sc. and his Masters degree in mathematics from the University of Wales Swansea (2004) and from the University of Mannheim (2005). In 2009, he obtained a Ph.D. in computer science from the Saarland University and the Max Planck Institut für Informatik. He was a postdoctoral researcher at the Computer Vision Laboratory, ETH Zurich, from 2009 until 2012 and senior research scientist at the Max Planck Institute for Intelligent Systems in Tübingen from 2012 until 2013. Since 2013, he is professor at the University of Bonn and head of the Computer Vision Group.

\end{IEEEbiography}




\end{document}